%% file: blessing.tex
\documentclass[final]{cvpr}
\usepackage{times}
\usepackage{epsfig}
\usepackage{graphicx}
\usepackage{amsmath}
\usepackage{amssymb}
\usepackage{multirow}
\usepackage{url}
\usepackage{times}  
\usepackage{helvet} 
\usepackage{courier}  
\usepackage[hyphens]{url}  
\usepackage{graphicx} 
\usepackage{caption} 
\usepackage{amsfonts}
\usepackage{xcolor}
\usepackage{booktabs}
\usepackage[ruled,linesnumbered]{algorithm2e}

\usepackage{algorithmic}
\usepackage{amsmath}
\usepackage[switch]{lineno}
\SetKwInput{KwData}{Input}
\SetKwInput{KwResult}{Output}

\definecolor{orange}{rgb}{1.0, 0.5, 0.0}
\definecolor{MyDarkGreen}{rgb}{0.02,0.6,0.02}

\usepackage[pagebackref=true,breaklinks=true,colorlinks,bookmarks=false]{hyperref}

\def\cvprPaperID{2266} 
\def\confYear{CVPR 2021}

\begin{document}

\title{The Blessings of Unlabeled Background in Untrimmed Videos}
\author{ Yuan Liu$^1$ \quad
Jingyuan Chen$^1$\quad
 Zhenfang Chen$^2$ \quad\\
 Bing Deng$^1$\quad
 Jianqiang Huang$^1$ \quad
 Hanwang Zhang$^3$ \\
$^1$Alibaba Group\quad $^2$The University of Hong Kong \quad $^3$Nanyang Technological University\\ 
\tt\small lhy19930911@gmail.com\qquad\qquad jingyuanchen91@gmail.com\qquad \qquad zfchen@cs.hku.hk\\
\tt\small dengbing.db@alibaba-inc.com\qquad jianqiang.jqh@gmail.com \qquad  hanwangzhang@ntu.edu.sg
}

\maketitle
\input{text/abstract}
\input{text/intro}

\input{text/related}

\input{text/method1}

\input{text/method2}

\input{text/experiment}

\input{text/conclusion}
{\small
\bibliographystyle{ieee_fullname}
\bibliography{egbib.bib}
}
\clearpage
\appendix

\input{text/supp}

\end{document}

%% file: text/abstract.tex
\begin{abstract}
Weakly-supervised Temporal Action Localization (WTAL) aims to detect the action segments with only video-level action labels in training. The key challenge is how to distinguish the action of interest segments from the background, which is unlabelled even on the video-level. While previous works treat the background as ``curses'', we consider it as ``blessings''. Specifically, we first use causal analysis to point out that the common localization errors are due to the unobserved confounder that resides ubiquitously in visual recognition. Then, we propose a Temporal Smoothing PCA-based (TS-PCA) deconfounder, which exploits the unlabelled background to model an observed substitute for the unobserved confounder, to remove the confounding effect. Note that the proposed deconfounder is model-agnostic and non-intrusive, and hence can be applied in any WTAL method without model re-designs. Through extensive experiments on four state-of-the-art WTAL methods, we show that the deconfounder can improve all of them on the public datasets: THUMOS-14 and ActivityNet-1.3\footnote{Code is available at \url{https://github.com/liuyuancv/WTAL_blessing}}. 
\end{abstract}

%% file: text/intro.tex
\section{Introduction}
\label{sec:intro}
Temporal action localization aims to locate the start and end frames of actions of interest in untrimmed videos, \eg, \textsc{CliffDive} and \textsc{PassBall}. Weakly-supervised Temporal Action Localization (WTAL)~\cite{stpn,autoloc,untrimmednet} can train such a localizer by using only the video-level annotations like ``the video contains \textsc{CliffDive}'', without specifying its start and end frames. Therefore, WTAL is especially valuable in the era of video-sharing social networking service~\cite{chen2016micro}, where billions of videos have only video-level user-generated tags.

 \begin{figure}[htb]
\centering
\includegraphics[width=8cm]{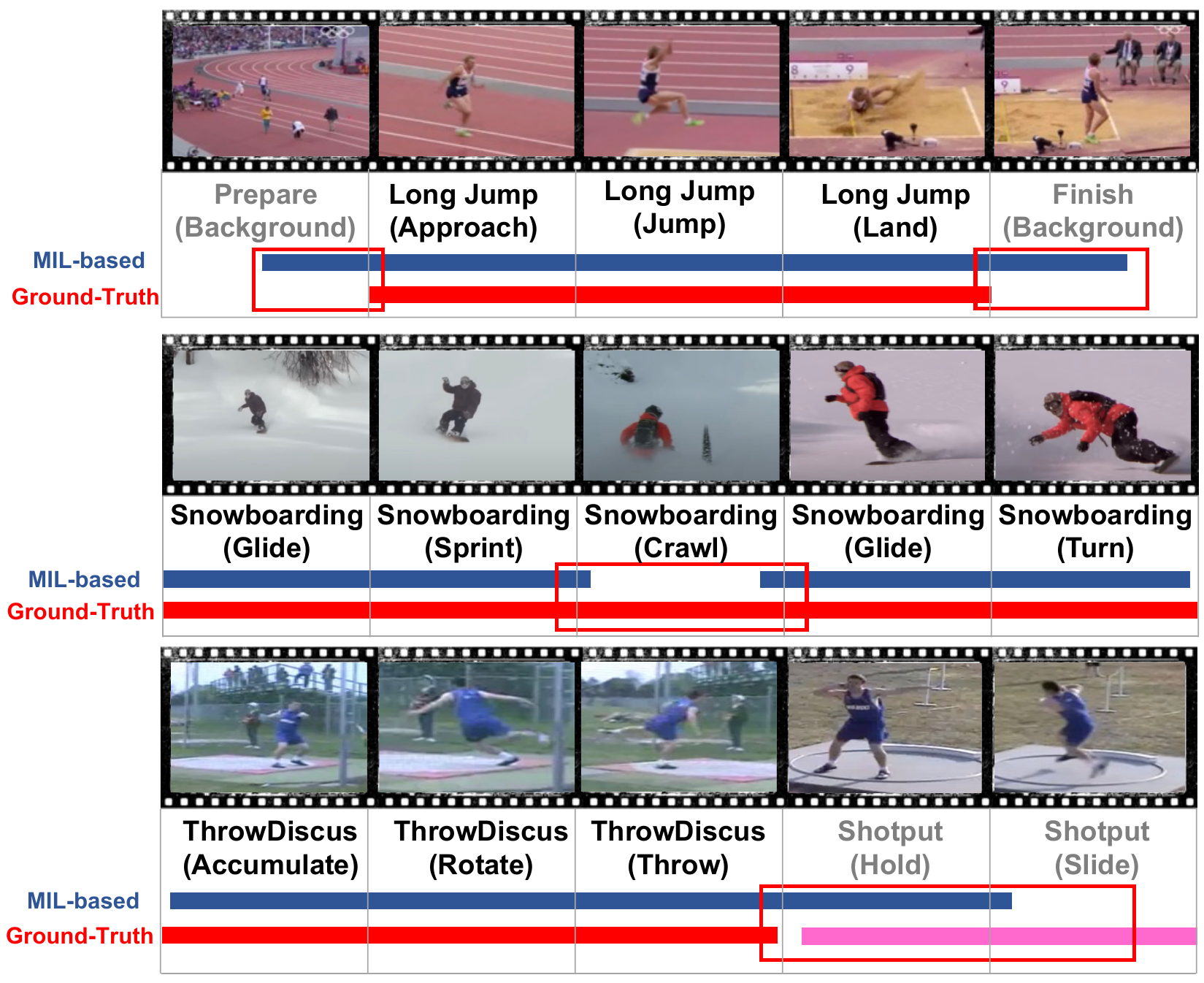}

\caption{Three types of common errors in existing MIL-based WTAL models. Top to Bottom: over-completeness, Incompleteness, Confusion. We argue that the errors are due to the unobserved confounders that affects the video (including both foreground and background) and the label simultaneously.
}
\label{sample}
\vspace{-1em}
\end{figure}

Not surprisingly, WTAL has lower performances than its fully-supervised counterpart~\cite{re_faster,bsn,mgg}. As shown in Figure~\ref{sample}, any WTAL method would suffer from the three types of localization errors due to the weak supervision: 1) \emph{Over-completeness}: the ground-truth is a sub-sequence of the localization, which contains additional background frames (Top). 2) \emph{Incompleteness}: the localization only selects the discriminative sub-sequence of the ground-truth, \ie, misclassifying foreground into background (Mid). 3) \emph{Class Confusion}: the localization misclassifies different actions (Bottom). Prior studies would always blame these errors for the unlabeled background segments, because the reasons are intuitively sound. For example, if we could model the background as a special \textsc{Background} class, error 1 and error 2  would be mitigated since WTAL knows what visual cues are background~\cite{basnet,multi_branch}; if we could exploit the difference between the background and foreground, each action class model would be more discriminative, thanks to the training on more negative background samples~\cite{wum,stpn,w_talc}---all the three errors are addressed.

However, it is impossible to fundamentally resolve the ``background'' issue by only using video-level supervision. The reasons are two-fold. First, as there is neither foreground nor background video-level labels, even if we set up a \textsc{Background} class, the \textsc{Background} model will be eventually reduced to a simple prior, such as the foreground is sparsely distributed in the background~\cite{stpn,dgam}. Second, learning more discriminative features between background and foreground is no easier than WTAL \textit{per se}, because if we have had such features, videos would be trimmed into segments of foreground actions, which are no longer weakly-supervised anymore. As a result, all such methods essentially resort to bootstrapping from another trivial prior~\cite{3c_net,w_talc}: segments of the same action should look similar, which is however not always the case for complex actions (\eg, \textsc{ThrowDiscus} in Figure~\ref{sample}). 

In this paper, we propose a \emph{causal inference} framework~\cite{pearl,rubin} to identify the true cause---which is indeed not the background---of the WTAL deficiency, and then how to remedy it by eliminating the cause. To begin with, we help you to find out the possible confounders in Figure~\ref{sample}. In causal terms, a \emph{confounder} is the common cause of many resultant effects, which are observed to be correlated even if they share no causation. For example, the context ``athletic track'' may be the confounder between the video and its video-level label \textsc{LongJump}---even if the ``Prepare''/``Finish'' background is not the action, it is inevitably correlated with the foreground and the label, misleading WTAL. The contextual confounder for the \textsc{ThrowDiscuss} and \textsc{Shotput} confusion can be found similarly. As another example, the object ``leg on snowboard'' may be the confounder for the \textsc{Snowboarding} video, where if a segment has no such object, it is wrongly detected as negative. 

It is well-known that such confounding bias can be \emph{never} removed by using only the observational data correlation~\cite{pearl2009causality}, \eg, the action likelihood estimated from a Multiple Instance Learning (MIL) model~\cite{3c_net}. Instead, we should perform \emph{causal intervention} to remove the confounding effect. Ideally, if we can observe all the possible confounders, we can estimate the average effect of the video evenly associated with each confounder~\cite{backdoor, zhangdong}, mimicking the physical intervention such as ``recording'' the videos of \textsc{LongJump} everywhere. Unfortunately, the above observed-confounder assumption is not valid in general, because the confounder is too elusive to be defined in different actions that have complex visual cues, \eg, it can be the scene context in \textsc{LongJump} or the object in \textsc{Snowboarding}. Fortunately, we can use another way around: Deconfounder~\cite{blessing}, which is a general theory that uses latent variables, who can generate the observed data, to be a good substitute for the confounders. 

To implement the deconfounder for WTAL, we propose an unsupervised
Temporal Smoothing PCA (TS-PCA) model whose base can reconstruct the whole video dataset, whose majority consists of the unlabelled background segments (Section~\ref{sec:pca}). Therefore, our punchline is that the background is indeed not a curse but blessings for WTAL. It is worth noting that the TS-PCA deconfounder is specially designed that each projection directly contributes to the label prediction logits. This has two crucial benefits: 1) Its implementation is model-agnostic, \ie, any existing WTAL model can seamlessly incorporate it in a plug-and-play fashion (Section~\ref{sec:sub_confounder}). 2) The TS-PCA projection can be used as a foreground/background score function that further enhances the prediction (Section~\ref{sec:pca}).
The deconfounder model is applied to four state-of-the-art WTAL methods with public codes on THUMOS-14 and ActivityNet-1.3 datasets, where a significant improvement is observed and the deconfounded WUM~\cite{wum} establishes a new state-of-the-art performance (Section~\ref{sec:exp}).

%% file: text/related.tex
\section{Related Work}

\noindent\textbf{Weakly-supervised Temporal Action Localization}~\cite{prototypical,wum,3c_net,w_talc}. UntrimmedNets~\cite{untrimmednet} locates action instances by selecting relevant segments in soft and hard ways.
Autoloc~\cite{autoloc} directly predicts the temporal boundary of each action instance with an outer-inner-contrastive loss.
BasNet~\cite{basnet} proposes an auxiliary class representing background to suppress the activation from background frames.
However, these methods have localization errors: over-completeness and incompleteness.
Many approaches have been proposed to tackle the issues by erasing salient features~\cite{hide_and_seek,zhong2018step}, imposing diversity loss~\cite{multi_branch}, and employing marginalized average aggregation strategy~\cite{maan}.
Nevertheless, these methods ignore the challenging action-context confusion issue caused by the absence of frame-wise label.
DGAM~\cite{dgam} builds a generative model for the frame representation, helpful for the action-context separation. However, all components of DGAM are coupled together, which cannot be utilized in a plug-and-play fashion.
In this paper, we propose an unsupervised TS-PCA deconfounder to solve the WTAL problem by contributing a principled answer based on causal inference. 

\noindent\textbf{Causal Inference.}
Causal inference can be used to remove the spurious bias~\cite{bareinboim2012controlling} and disentangle the desired model effects~\cite{besserve2018counterfactuals} in domain-specific applications by pursuing the causal effect~\cite{pearl2016causal,rubin2019essential}. Recently, 
causal inference has been introduced and widely used in various computer vision tasks~\cite{niu2020counterfactual,qi2020two,zhangdong}, including image classification~\cite{chalupka2014visual}, few-shot learning~\cite{yue2020interventional}, and semantic segmentation~\cite{zhangdong}. In our work, as the confounder is unobserved, those deconfounding methods based on the observed-confounder assumption is no longer valid. To this end, we propose to derive a substitute confounder from a fitted data generation model~\cite{blessing}.

%% file: text/method1.tex
\section{Approach}
In this section, we discuss how to derive our deconfounding techniques from the causal effect perspective and how to implement it into the prevailing WTAL framework.

\subsection{Action Localization as Causal Effects}
\noindent\textbf{Task Definition.}
Weakly-supervised temporal action localization (WTAL) aims to predict a set of action instances given an untrimmed video.
Suppose we have $N$ training videos $\{\mathbf{x}_i \}_{i=1}^N$ with their video-level labels $\{\mathbf{y}_i \}_{i=1}^N$, 
where $\mathbf{y}_i$ is a $F$-dimensional multi-hot vector representing the presence of $F$ action categories and 
$\mathbf{x}_i\in  \mathbb{R}^{T\times D}$ is the extracted $D$-dimensional video features of $T$ segments.

\begin{figure}[t]
\centering
\includegraphics[width=8.5cm]{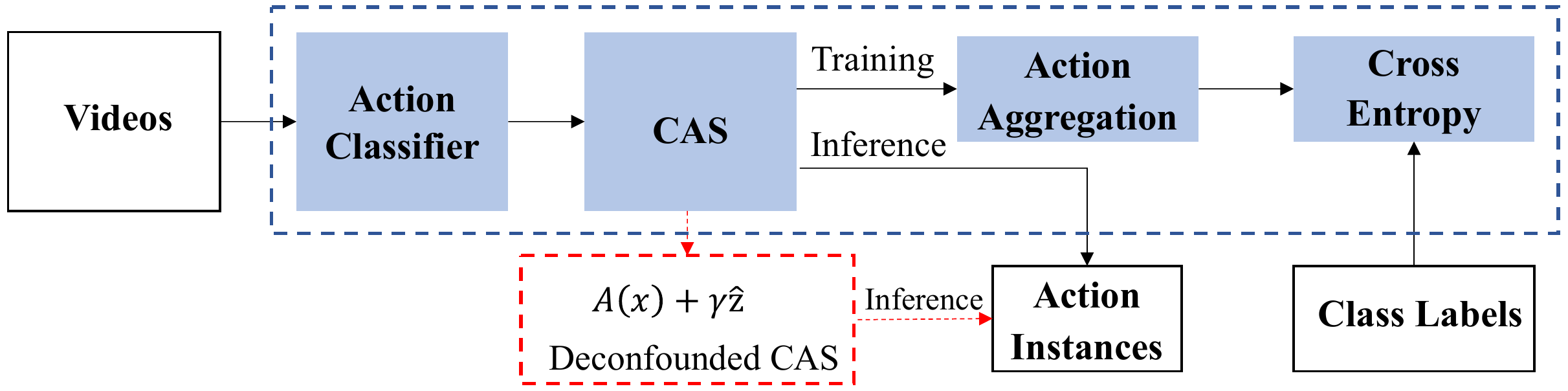}
\caption{The blue dashed box denotes the prevailing pipeline for WTAL. Our contribution is to calibrate CAS with the deconfounder during the inference process, which is shown in the red dashed box.}
\label{old_pipeline}
\end{figure}

The prevailing pipeline for training WTAL is illustrated in the blue dashed box of
Figure~\ref{old_pipeline}.
Note the whole framework is trained with only video-level action labels. First, video $\mathbf{x}_i$ goes through an action classifier to generate segment-wise scores for each action category, namely, Class Activation Sequences~(CASs), which are denoted as $A(\mathbf{x}_i)\in\mathbb{R}^{T\times F}$. The CAS score reflects the probability of each segment whether it belongs to a specific action category. 
Then, CASs are aggregated to produce the video-level class scores  $\mathbf{y}_i=\varphi (A(\mathbf{x}_i))\in\mathbb{R}^F$. Specifically, $\varphi(\cdot)$ is usually a rule-based function and the most widely used one is top-$k$ mean method~\cite{basnet,3c_net,w_talc}, which averages the top-$k$ scores of all segments for each action category, followed by a softmax operation to derive the video-level action probabilities. Therefore, the video-level classification model can be seen as a proxy objective for CAS learning, with a binary cross-entropy loss for each class as:
\begin{equation}
\label{eq_cls_loss}
    \mathcal{L}_{cls} = -\frac{1}{N}\sum_{i=1}^{N}\sum_{f=1}^{F}y_{i,f}log(\varphi (A(\mathbf{x}_{i}))_f).
\end{equation}
During inference, action instances are localized by thresholding $A(\mathbf{x}_i)$ for segment-wise action intervals.
However, as we discussed in Section~\ref{sec:intro}, it is essentially ill-posed to infer $A(\mathbf{x}_i)$ from only video-level labels via Eq.~(\ref{eq_cls_loss}) due to the unobserved confounder. In the following discussion, when the context is clear, we omit the subscript of each sample for simplicity.

\noindent\textbf{Causal Effect Analysis.} We slightly abuse the notation $A$ to denote that there exists an ``ideal'' CAS $A$ for the ground-truth video-level action label $\mathbf{y}$. In causal inference, $A$ is also known as the \emph{potential outcome}~\cite{rubin}, and the challenge is how to use and only use the training data to estimate it. Denote the  conditional Monte Carlo approximations~\cite{burt1971conditional} of $A$ as  $\mathbb{E}[A|X=\mathbf{x}]$.  However, as shown in Figure~\ref{figdeconfounder}, when there is confounder $C$, it will affect both the input video features $\mathbf{x}$ and CAS $A$, \ie, $\mathbf{x}$ may correlate with the observed label $\mathbf{y}$, even if $\mathbf{x}$ is not the true cause (Recall the localization errors illustrated in Figure~\ref{sample}). Therefore, the Monte Carlo estimation $\mathbb{E}[A|X=\mathbf{x}]$ is biased,
\begin{equation}
    \mathbb{E}[A|X=\mathbf{x}]\neq A.
\end{equation}

Suppose we could observe and measure all the possible confounders $C$ and associate them to each data sample,
\begin{equation}
    \mathbb{E}_c\left[\mathbb{E}[A|X = \mathbf{x}, C = c]\right]=A,
    \label{eq:aug}
\end{equation}
which is also known as the backdoor adjustment~\cite{pearl2009causality}. Unfortunately, the above assumption of observed confounder is not valid in general because the confounder is ever-elusive to be defined in different actions of complex visual cues. Next, we propose a deconfounder method to remove the confounding effect.

%% file: text/method2.tex
\subsection{Deconfounded WTAL}
We now develop the deconfounder~\cite{blessing} for WTAL task, which is a general algorithm that uses latent variables which generate the observed data, as a good substitute for the unobserved confounders (Section~\ref{sec:sub_confounder}). Note that the substitute is obtained from an unsupervised generative model learned from all videos---regardless of foreground or background, labeled or unlabeled---the blessings of unlabeled background (Section~\ref{sec:pca}).

\begin{figure}[t]
\centering
\includegraphics[width=7cm]{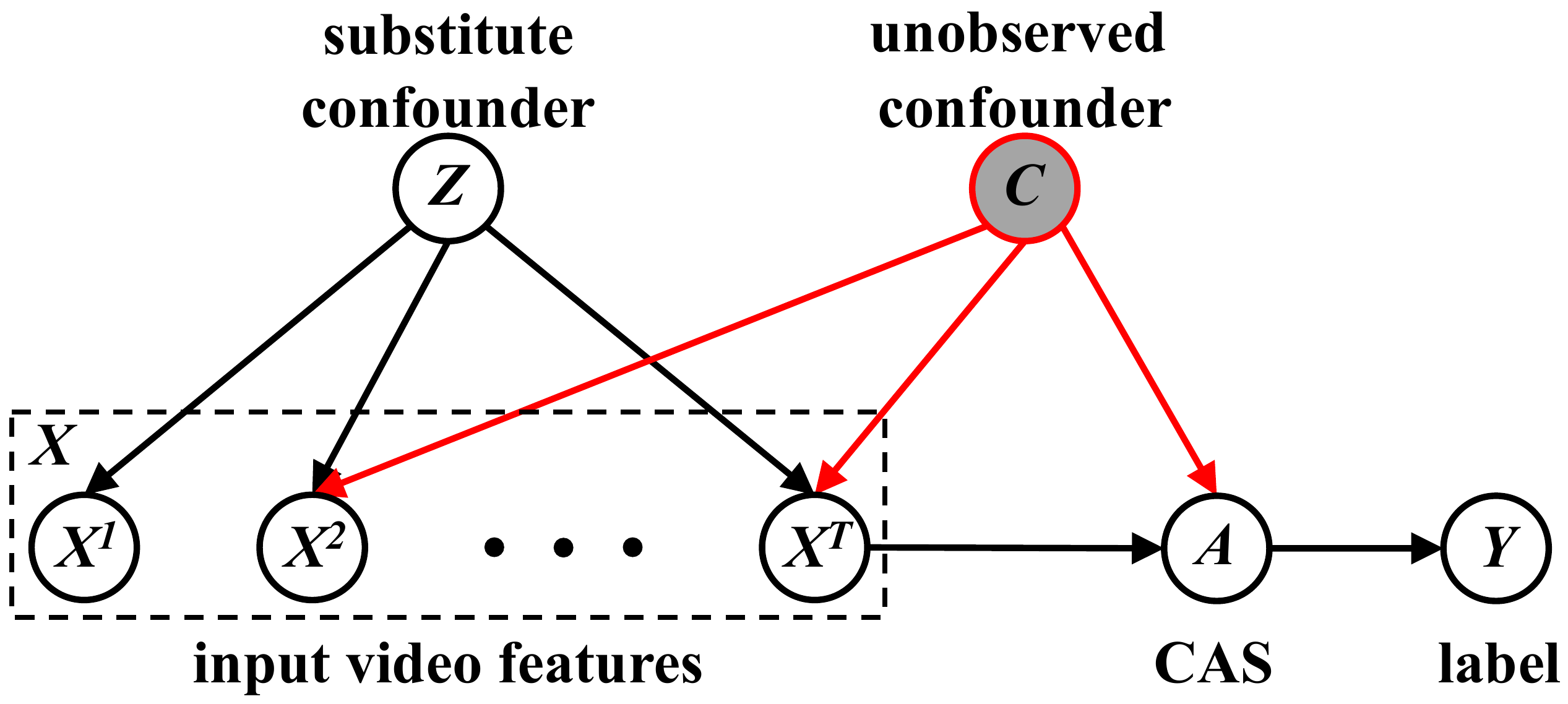}
\caption{The unobserved confounders $C$ are correlated to the input video $X$ and CAS $A$. The substitute confounder $Z$ captures the distribution of all the input video features. Therefore, all the confounders are captured with the substitute confounders.
}
\label{figdeconfounder}
\vspace{-1.5em}
\end{figure}
\subsubsection{Deconfounded CAS Function}
\label{sec:sub_confounder}
The deconfounder theory infers a substitute confounder to perform causal inference.
As shown in Figure~\ref{figdeconfounder}, we assume that there exists a substitute confounder $Z$ which could generate the distribution of the input video features $\mathbf{x}=\{\mathbf{x}^1,\mathbf{x}^2,\cdots,\mathbf{x}^T\}$. Therefore, all the segment-level features are conditionally independent given $\mathbf{z}$:
\begin{equation}
    \label{eq_dependence}
    P(\mathbf{x}^1,...,\mathbf{x}^T\mid Z=\mathbf{z}) = \prod_{t=1}^{T}P(\mathbf{x}^t\mid  Z=\mathbf{z}).
\end{equation}
Now we show that $\mathbf{z}$ is a good substitute which can capture \emph{all} the unobserved confounders $C$ by contradiction. Suppose that there exists an unobserved confounder $c$, which affects multiple input video features within $\mathbf{x}$ and segment-level labels $A$.
Then, $\mathbf{x}$ would be dependent, even conditional on $\mathbf{z}$, due to the impact of $c$. This dependence leads to a contradiction that Eq.~(\ref{eq_dependence}) does not hold.

Given the substitute confounder $\mathbf{z}$ who can generate $\mathbf{x}$, we can replace the expectation over $C$ with a single $\mathbf{z}$ in Eq.~(\ref{eq:aug}), thanks to the (weak) ignorability~\cite{imai2004causal,rosenbaum1983central}:
 \begin{equation}
     A=\mathbb{E}[A|X=\mathbf{x},Z=\mathbf{z}] := 
\mathbb{E}[G(\mathbf{x},\mathbf{z})],
\label{eq:5}
 \end{equation}
where $G(\cdot)$ is the outcome model to estimate $A$. In particular, if we denote $\hat{\mathbf{z}}$ as a deterministic inference result (\ie, decoded) from the generative model $\hat{\mathbf{z}} \sim P(\mathbf{z}|\mathbf{x})$, the expectation in Eq.~\eqref{eq:5} will be collapsed to:
 \begin{equation}
     A=
\mathbb{E}[G(\mathbf{x},\hat{\mathbf{z}})] = G(\mathbf{x},\hat{\mathbf{z}}).
 \end{equation}
We propose to use a decoupled addition model to be the overall deconfounded CAS function $\widehat{A}(\mathbf{x})$:
\begin{equation}
 \label{eq_linear}
    \widehat{A}(\mathbf{x}) := G(\mathbf{x},\hat{\mathbf{z}})=A(\mathbf{x})+\gamma \hat{\mathbf{z}},
\end{equation}
where $A(\cdot)$ is the conventional CAS function in any WTAL model, 
and $\gamma$ is a trade-off parameter. It is worth noting that the deconfounded CAS $\widehat{A}(\mathbf{x})$ is model-agnostic---all we need to achieve deconfounded WTAL is just to calibrate any trained and conventional $A(\mathbf{x})$ by adding $\gamma \hat{\mathbf{z}}$ as shown in the dashed red box of Figure~\ref{old_pipeline}. Before we detail how to obtain $\hat{\mathbf{z}}$ in Section~\ref{sec:pca}, we first outline the deconfounded WTAL based on the deconfounded CAS function in Eq.~\eqref{eq_linear}.

Algorithm~\ref{pipeline} illustrates the overview of the deconfounded WTAL. The inputs are training videos with only video-level class labels and the output is the deconfounded CAS scores $\widehat{A}(\mathbf{x})$. Given any WTAL model $A(\mathbf{x})$ and the generative model which is to encode $\hat{\mathbf{z}}$, the two models are trained separately.
With such decoupled training scheme, any existing WTAL model can be seamlessly incorporated with  $\hat{\mathbf{z}}$ in a plug-and-play fashion. Specifically,
$A(\mathbf{x})$ is typically trained with the loss function in Eq.~(\ref{eq_cls_loss}) and the generative model is trained with $\mathcal{L}_{PCA}$ detailed in Eq.~(\ref{eq_project}). During the inference process,
based on the trained $A(\mathbf{x})$  and the generated $\hat{\mathbf{z}}$, Eq.~(\ref{eq_linear}) is adopted to generate the deconfounded CAS scores, which is also shown in Figure~\ref{old_pipeline}. 
Action instances are inferred by thresholding $\widehat{A}(\mathbf{x}_i)$, which is implemented with existing methods~\cite{basnet,3c_net,w_talc}.

\begin{algorithm}[t]
\caption{WTAL with the Deconfounder}\label{pipeline}

\KwData{input videos and action categories $\{\mathbf{x}, \mathbf{y}\}$.}
\KwResult{Deconfounded CAS $\widehat{A}(\mathbf{x})$ for each $\mathbf{x}$.}
\begin{algorithmic}[1]
\REPEAT

\STATE {Training any WTAL model with $\mathcal{L}_{cls}$ in Eq.~(\ref{eq_cls_loss}).}
 
\STATE {Training TS-PCA with $\mathcal{L}_{PCA}$ in Eq.~(\ref{eq_project}).
} 

\UNTIL {Converge.}


\FOR{\text{each video} $\mathbf{x}$}
\STATE Generating $A(\mathbf{x})$ with the trained WTAL model.
\STATE Generating $\hat{\mathbf{z}}$ by projecting $\mathbf{x}$ with the trained projectors of TS-PCA.
\STATE Generating final result $\hat{A}(\mathbf{x})$ according to Eq.~(\ref{eq_linear}).
\ENDFOR
\end{algorithmic}
\end{algorithm}

\subsubsection{Temporal Smoothing PCA (TS-PCA)}
\label{sec:pca}
To be a good substitute confounder, $\hat{\mathbf{z}}$ needs to satisfy: 1) it captures the joint distribution of input features $\mathbf{x}$; and 2) $G(\mathbf{x},\hat{\mathbf{z}})$ is model-agnostic, which requires $\hat{\mathbf{z}}$ to generate CAS scores to be seamlessly fused with any WTAL model. Subject to these constraints, we propose to use a temporal smoothing PCA (TS-PCA) as a simple yet effective generative model for encoding the substitute confounder $\hat{\mathbf{z}}$. The model directly projects the features into a hidden factor $\mathbf{\hat{z}}$, which can be discriminative and naturally reflects the background or foreground. Without such unsupervised technique, it will require to train the projection together with $A(\mathbf{x})$, which ruins the decoupled training requirement of deconfounder.

Generally, TS-PCA learns $L$ feature projectors $\mathbf{P}=\{\mathbf{p}_l\}_{l=1}^L$ to maximize the variance of the projected features $\mathbf{x}$. Given the training video set $\{\mathbf{x}\}$, the overall objective function of TS-PCA is:
\begin{equation}
\label{eq_project}
    \mathcal{L}_{PCA} = \mathcal{L}_{pro} + \lambda \mathcal{L}_{recon} + \beta \mathcal{L}_{smooth},
\end{equation}
where $\lambda$ and $\beta$ are parameters to control the relative contributions. $\mathcal{L}_{pro}$ denotes the projection loss, which is the basis of PCA and composed of a variance and an orthogonalization term  as:
\begin{equation}
\label{eq:project}
\small{
\begin{aligned}
\mathcal{L}_{pro} = & -\sum_{t}\sum_{l}\left ( \mathbf{p}_l^{\intercal}\mathbf{x}^{t}-\frac{\sum_{m=1}^{T} \mathbf{p}_l^{\intercal}\mathbf{x}^{m}}{T} \right )^2 \\
&+\frac{1}{L^2}\left \| PP^{\intercal}-I \right \|_F.
\end{aligned}
}
\end{equation}

Moreover, with the feature projectors, 
we can approximately reconstruct the original features by mapping it back with $\mathbf{p}_l$. The reconstruction loss $\mathcal{L}_{recon}$ is as follows:
\begin{equation}
     \mathcal{L}_{recon} =\sum_{t} \left \|\sum_{l}(\mathbf{p}_l^{\intercal}\mathbf{x}^{t})\mathbf{p}_l -\mathbf{x}^{t} \right \| ^2.
\end{equation}

Motivated by the fact that nearby segments tend to have similar visual semantics, we propose a temporal smoothing loss to penalize the feature projection differences of consecutive segments:
\begin{equation}
    \label{loss:smooth}
    \mathcal{L}_{smooth} = \sum_{t}\sum_{l}\left (\mathbf{p}_l^{\intercal}\mathbf{x}^{t+1}- \mathbf{p}_l^{\intercal}\mathbf{x}^{t} \right )^2.
\end{equation}

For each video, we can calculate the substitute confounder $\mathbf{\hat{z}}$ using the TS-PCA above:
\begin{equation}
    \hat{\mathbf{z}} = \frac{1}{L}\sum_{l=1}^{L}\mathbf{p}_l^{\intercal}\mathbf{x},
\end{equation}
where $\hat{\mathbf{z}}\in\mathbb{R}^{T}$ 
is computed by projecting features with the trained projectors. We broadcast $\hat{\mathbf{z}}$ to $\mathbb{R}^{T\times F}$, which could be added to $\widehat{A}(\mathbf{x})$ shown in Eq.~\eqref{eq_linear}.

%% file: text/experiment.tex
\begin{table*}
\footnotesize \centering \caption{\label{table_thumos_pro}The deconfounder strategy is applied to different` state-of-the-art WTAL models on THUMOS-14 dataset. ``+TP'' denotes the TS-PCA deconfounder strategy is deployed. mAP values with IoU threshold $0.1$:$0.1$:$0.9$ are reported. $\dag$ indicates the result
is our re-implementation with the publicly available code.
}
\vspace{-1em}
\begin{tabular}{c|c|c|c|c|c|c|c|c|c}
\toprule Method &0.1 & 0.2&0.3 &0.4 &0.5&0.6 &0.7 &0.8 &0.9\\\midrule
 STPN~\cite{stpn} &52.0&44.7&35.5&25.8&16.9&9.9&4.3&1.2&0.1\\
STPN+TP&53.2&46.1&36.8&27.4&18.0&11.2&5.8&2.2&0.3\\
 Abs. Improve&\textcolor{blue}{+1.2}&\textcolor{blue}{+1.4}&\textcolor{blue}{+1.3}&\textcolor{blue}{+1.6}&\textcolor{blue}{+1.1}&\textcolor{blue}{+1.3}&\textcolor{blue}{+1.5}&\textcolor{blue}{+1.0}&\textcolor{blue}{+0.2}\\
\midrule
A2CL-PT~\cite{min2020adversarial}&61.2&56.1&48.1&39.0&30.1&19.2&10.6&4.8&1.0\\
A2CL-PT+TP&63.1&58.1&49.6&40.4&31.5&19.3&10.9&5.1&1.0\\
 Abs. Improve&\textcolor{blue}{+1.9}&\textcolor{blue}{+2.0}&\textcolor{blue}{+1.5}&\textcolor{blue}{+1.4}&\textcolor{blue}{+1.4}&\textcolor{blue}{+0.1}&\textcolor{blue}{+0.3}&\textcolor{blue}{+0.3}&\textcolor{red}{+0.0}\\
\midrule
 BasNet~\cite{basnet}&58.2&52.3&44.6&36.0&27.0&18.6&10.4&3.9&0.5\\
BasNet+TP&60.3&53.6&45.7&37.1&28.5&19.8&11.4&3.7&0.5\\
 Abs. Improve &\textcolor{blue}{+2.1}&\textcolor{blue}{+1.3}&\textcolor{blue}{+1.1}&\textcolor{blue}{+1.1}&\textcolor{blue}{+1.5}&\textcolor{blue}{+1.2}&\textcolor{blue}{1.0}&\textcolor{red}{-0.2}&\textcolor{red}{+0.0}\\
\midrule
WUM~\cite{wum} &67.5&61.2&52.3&43.4&33.7&22.9&12.1&3.9&0.5\\
WUM+TP&67.6&61.1&53.4&43.4&34.3&24.7&13.7&4.5&0.7\\
Abs. Improve &\textcolor{blue}{+0.1}&\textcolor{red}{-0.1}&\textcolor{blue}{+1.1}&\textcolor{red}{+0.0}&\textcolor{blue}{+0.6}&\textcolor{blue}{+1.8}&\textcolor{blue}{+1.6}&\textcolor{blue}{+0.6}&\textcolor{blue}{+0.2}\\
WUM$\dag$~\cite{wum} &64.9&59.2&50.8&41.8&32.6&21.0&10.6&3.1&0.3\\
 WUM$\dag$+TP &66.0&60.3&52.4&43.5&34.6&23.7&12.6&4.3&0.6\\
 Abs. Improve&\textcolor{blue}{+1.1}&\textcolor{blue}{+1.1}&\textcolor{blue}{+1.6}&\textcolor{blue}{+1.7}&\textcolor{blue}{+2.0}&\textcolor{blue}{+2.7}&\textcolor{blue}{+2.0}&\textcolor{blue}{+1.2}&\textcolor{blue}{+0.3}\\
\bottomrule
\end{tabular}
\vspace{-1em}
\end{table*}

\begin{table}
\footnotesize \centering \caption{\label{table_activitynet_pro}The deconfounder applied to state-of-the-art WTAL models on ActivityNet-1.3 dataset. 
mAP values with IoU threshold 0.5, 0.75, and 0.95 are reported. The column AVG denotes the average mAP values at IoU thresholds 0.5:0.05:0.95.  $*$ indicates that the reported results are not present in the original paper.}
\vspace{-1em}
\begin{tabular}{c|c|c|c|c}
\toprule Method &0.5 & 0.75&0.95 &AVG\\\midrule
 STPN ~\cite{stpn} &29.3&16.9&2.6&15.9$^\ast$\\
STPN+TP&31.3&18.0&3.6&17.7\\
 Abs. Improve&\textcolor{blue}{+2.0}&\textcolor{blue}{+1.1}&\textcolor{blue}{+1.0}&\textcolor{blue}{+1.8}\\
\midrule
A2CL-PT~\cite{min2020adversarial}&36.8&22.0&5.2&22.5\\
 A2CL-PT+TP&37.4&23.5&5.9&23.7\\
 Abs. Improve&\textcolor{blue}{+0.6}&\textcolor{blue}{+1.5}&\textcolor{blue}{+0.7}&\textcolor{blue}{+1.2}\\
\bottomrule
\end{tabular}
\vspace{-1em}
\end{table}

\begin{table*}
\footnotesize \centering \caption{\label{sota_thu}Comparisons with the state-of-the-art methods on the testing set of THUMOS-14. 
We denote fully-supervised and weakly-supervised as Full and Weak, respectively.
+TP represents the proposed deconfounder.
mAP values with IoU threshold $0.1$:$0.1$:$0.9$ are reported.}
\label{tab:tab2}
\vspace{-1em}
\begin{tabular}{c|c|c|c|c|c|c|c|c|c|c}
\toprule Supervision &Method &0.1 & 0.2&0.3 &0.4 &0.5&0.6 &0.7 &0.8 &0.9\\\midrule
\multirow{4}{*}{Full} 
& R-C3D~\cite{r-c3d}     &54.5&51.5&44.8&35.6&28.9&-&-&-&-\\
& SSN~\cite{ssn}          &66.0&59.4&51.9&41.0&29.8&-&-&-&-\\
& MGG~\cite{mgg}          &-&-&53.9&46.8&37.4&29.5&21.3&-&-\\
& GTAN~\cite{gtan}        &69.1&63.7&57.8&47.2&38.8&-&-&-&-\\
& P-GCN~\cite{pgcn}       &69.5&67.8&63.6&57.8&49.1&-&-&-&-\\
& G-TAD~\cite{gtad}      &-&-&66.4&60.4&51.6&37.6&22.9&-&-\\

\midrule
\multirow{14}{*}{Weak}
&UntrimmedNets~\cite{untrimmednet} &44.4&37.7&28.2&21.1&13.7&-&-&-&-\\
&Hide-and-seek~\cite{hide_and_seek} &36.4&27.8&19.5&12.7&6.8&-&-&-&-\\
&AutoLoc~\cite{autoloc}&-&-&35.8&29.0&21.2&13.4&5.8&-&-\\
& MAAN~\cite{maan} &59.8&50.8&41.1&30.6&20.3&12.0&6.9&2.6&0.2\\
& W-TALC~\cite{w_talc}&55.2&49.6&40.1&31.1&22.8&-&7.6&-&-\\
& DGAM~\cite{dgam}&60.0&54.2&46.8&38.2&28.8&19.8&11.4&3.6&0.4\\
& TSCN~\cite{zhai2020two} &63.4&57.6&47.8&37.7&28.7&19.4&10.2&3.9&0.7\\
 &A2CL-PT~\cite{min2020adversarial} &61.2&56.1&48.1&39.0&30.1&19.2&10.6&4.8&\textbf{1.0}\\
 &WUM~\cite{wum} &67.5&61.2&52.3&43.4&33.7&22.9&12.1&3.9&0.5\\
 & WUM+TP (Ours)&\textbf{67.6}&61.1&\textbf{53.4}&\textbf{43.4}&\textbf{34.3}&\textbf{24.7}&\textbf{13.7}&\textbf{4.5}&0.7\\
\bottomrule
\end{tabular}
\vspace{-1em}
\end{table*}

\section{Experiments}\label{sec:exp}

Following previous works~\cite{basnet,stpn}, we conduct experiments on THUMOS-14 and ActivityNet-1.3 benchmarks. We show the effectiveness of
 the TS-PCA  deconfounder on different baselines in Section~\ref{Generalization}. We then compare the proposed method with the state-of-the-art models in Section~\ref{exp:stoa}. We also validate the components in TS-PCA with detailed ablation studies in Section~\ref{exp:ablation}. Moreover, we show the false positive analysis and the qualitative results in Section~\ref{exp:fpa}
 and Section~\ref{exp:qr}, respectively.

\subsection{Settings}
\noindent\textbf{Datasets.}
We evaluate the proposed deconfounder strategy on two standard datasets for WTAL, namely THUMOS-14~\cite{thumos} and ActivityNet-1.3~\cite{activitynet}. \textbf{THUMOS-14} contains $200$ videos and $212$ videos with $20$ action classes in the validation and test sets, respectively.
Each untrimmed video contains at least one action category.
Following the standard practice of previous works~\cite{basnet,stpn}, we train the network on the validation set and evaluate the trained model on the test set.
\textbf{ActivityNet-1.3} is a large-scale video benchmark for temporal action localization, which consists of
$19,994$ videos with $200$ classes annotated, with $50$\% for training, $25$\% for validation, and the rest $25$\% for testing.
As in literature~\cite{w_talc,maan},
we train our model on the training set and perform evaluations on the validation set. 

\noindent\textbf{Evaluation Metrics.}
Following the standard evaluation metrics, we report mean average precision (mAP) at several different levels of intersection over union (IoU) thresholds. 
We use the official benchmarking code provided by \cite{activitynet} to evaluate our model on THUMOS-14 and ActivityNet-1.3.

\noindent\textbf{Baseline Models.} To demonstrate the generalizability of the proposed deconfounder strategy, we integrated it on four popular WTAL models including STPN~\cite{stpn}, A2CL-PT~\cite{min2020adversarial}, BasNet~\cite{basnet} and WUM~\cite{wum} with available source code. 
The four models utilize different strategies to improve the performance, serving as good baselines for demonstrating TS-PCA's generalizability for different WTAL models.
For a fair comparison, we follow the same settings as reported in the official codes of STPN, A2CL-PT, BasNet and WUM.

\noindent\textbf{Implementation Details.}
I3D networks~\cite{i3d} are used to extract segment features which take segments with 16 frames as input.
I3D networks are pre-trained on Kinetics~\cite{i3d}.
Note that the feature extractors are not finetuned during training.

\begin{table}
\footnotesize \centering \caption{\label{wtal_activitynet}Comparisons with the state-of-the-art methods on the validation set of ActivityNet-1.3.
We denote fully-supervised and weakly-supervised as Full and Weak, respectively.
+TP represents the proposed deconfounder. mAP values with IoU threshold of $0.5$, $0.75$, and $0.95$ are reported. The column AVG denotes the average mAP values with IoU thresholds $0.5$:$0.05$:$0.95$.
All methods employ I3D as the feature extractor. $*$ indicates that the reported results are not present in the original paper.}
\begin{tabular}{c|c|c|c|c|c}
\toprule Supervision &Method &0.5 & 0.75&0.95 &AVG \\
\midrule
\multirow{3}{*}{Full} & CDC~\cite{cdc}&45.3&26.0&0.2&23.8\\
& R-C3D~\cite{r-c3d}&26.8&-&-&12.7\\
& BSN~\cite{bsn}&52.5&33.5&8.9&34.3\\
& G-TAD~\cite{gtad}&50.4&34.6&9.0&34.1\\
\midrule
\multirow{7}{*}{Weak}& STPN~\cite{stpn} &29.3&16.9&2.6&15.9$^\ast$\\
& MAAN~\cite{maan} &33.7&21.9&5.5&-\\
 & Nguyen\textit{ et al.}~\cite{nguyen2019weakly}&36.4&19.2&2.9&-\\
&TSCN~\cite{zhai2020two}&35.3&21.4&5.3&21.7\\
& WUM~\cite{wum} &37.0&23.9&5.7&23.7\\
&A2CL-PT~\cite{min2020adversarial} &36.8&22.0&5.2&22.5\\
& A2CL-PT+TP(Ours)&\textbf{37.4}&23.5&\textbf{5.9}&\textbf{23.7}\\
\bottomrule
\end{tabular}
\vspace{-1em}
\end{table}

\subsection{Effectiveness on Different Baselines}
\label{Generalization}
To demonstrate the generalizability of TS-PCA, we deploy it with four WTAL models on the THUMOS-14 dataset, which are STPN~\cite{stpn}, A2CL-PT~\cite{min2020adversarial}, BasNet~\cite{basnet} and WUM~\cite{wum}. We also perform similar experiments with STPN and A2CL-PT~\cite{min2020adversarial} on the ActivityNet-1.3 dataset.
Experimental results on THUMOS-14 are shown in Table~\ref{table_thumos_pro}, where ``+TP'' denotes that the TS-PCA deconfounder is deployed.
We can observe that all four models with +TP obtain obvious improvements.
Specifically, there are averaged mAP improvement of $1.18$\% on STPN,  $0.99$\% on A2CL-PT, $1.01$\% on BasNet and $1.52$\% on WUM. As for ActivityNet-1.3 dataset, STPN and A2CL-PT with +TP also show similar performance improvement as shown in Table~\ref{table_activitynet_pro}.
Specifically, STPN+TP achieves an average mAP improvement of $1.8$\%, and A2CL-PT has an average improvement of $1.2$\%. 
These results show the effectiveness of TS-PCA deconfounder in calibrating CAS scores.
Note that WUM is a recently proposed \textbf{SOTA} method, where the absolute improvements of $1.52$\% on THUMOS-14 are significant. 
The above results demonstrate the effectiveness and generalizability of the deconfounder.

\subsection{Comparisons with State-of-the-Art Methods}
\label{exp:stoa}
We adopt the TS-PCA deconfounder on WUM as our model to compare with other methods on the testing set of THUMOS-14.
We first compare WUM+TP with several baseline models in both weakly-supervised and fully-supervised training manners.
The results are shown in Table~\ref{sota_thu}.
It can be observed that WUM+TP outperforms other weakly-supervised methods by a large margin, establishing new state-of-the-art performances.

Specifically, WUM+TP achieves an improvement of $5.3$\% on mAP at IoU=$0.3$ and $5.5$\% at IoU=$0.6$ compared with the second best model A2CL-PT.
These results further demonstrate the effectiveness of the TS-PCA deconfounder in calibrating CAS.
Moreover, our method can even achieve better results than some fully-supervised methods such as SSN~\cite{ssn} which is trained with temporal boundary annotations, showing the superiority of the proposed method.

Table~\ref{wtal_activitynet} shows the comparison results on ActivityNet-1.3, where we adopt the TS-PCA deconfounder on A2CL-PT\footnote{We choose A2CL-PT rather than WUM since WUM's source code and pretrain-ed model on ActivityNet-1.3 are not publicly available.}. It can be observed that A2CL-PT+TP outperforms other state-of-the-art weakly-supervised models.
Particularly,  A2CL-PT+TP surpasses WUM by $0.4$\% on mAP with IoU=$0.5$.
Moreover, A2CL-PT+TP achieves comparative performance with the fully supervised model CDC~\cite{cdc}.

\subsection{Ablation Studies}
\label{exp:ablation}
We further investigate the contributions of different components in the TS-PCA deconfounder in detail.
We conduct ablation studies on THUMOS-14 dataset.
Without loss of generality, we incorporate WUM with TS-PCA as the basic model to verify the effectiveness of different components. Specifically, the ablation studies aim to answer two questions. \textbf{Q1}: \textit{What's the contribution of each loss component?} \textbf{Q2}: \textit{How many feature projectors are needed to encode $\hat{\mathbf{z}}$?}

\vspace{-1em}
\subsubsection{Contributions of Different Loss Components}
To get a better understanding of the proposed TS-PCA deconfounder, we further evaluate the key components of $\mathcal{L}_{PCA}$.
\textbf{TS-PCA-R:} We discard the reconstruction loss, which is to encourage the reconstruction of video features based on the learned projectors, helpful for capturing the distribution of the input data.
\textbf{TS-PCA-S:} The smoothing loss is to penalize the feature projection differences of consecutive segments.  Here, we discard the smoothing loss to verify its contributions.

\begin{table}
\centering \small \caption{\label{ablation} 
Performance comparisons of TS-PCA and its variants (TS-PCA-R and TS-PCA-S) deployed on WUM 
on the testing set of THUMOS-14. mAP values with IoU threshold $0.3$:$0.1$:$0.7$ are reported.}
\vspace{-1em}
\begin{tabular}{c|c|c|c|c|c}
\toprule Method &0.3&0.4&0.5&0.6&0.7\\\midrule
TS-PCA-R &51.8&42.7&33.5&22.9&12.3\\
TS-PCA-S &51.9 &43.0 &33.7&23.1&12.5\\
\textbf{TS-PCA} &\textbf{52.3} &\textbf{43.4} &\textbf{34.0}&\textbf{23.6}&\textbf{12.8}\\
\bottomrule
\end{tabular}
\vspace{-1em}
\end{table}

As shown in Table~\ref{ablation}, WUM deployed with TS-PCA outperforms all its variants, namely  TS-PCA-R and TS-PCA-S, and achieves an average improvement of $0.58$\% and $0.38$\%, respectively.
Therefore, the reconstruction loss and the smoothing loss are both effective and necessary. 

\begin{table}
\centering \small \caption{\label{num_pro} 
With WUM~\cite{wum} as the baseline model, performance comparisons over different number of feature projectors on THUMOS-14 dataset. mAP values at IoU=$0.5$ is reported.}
\begin{tabular}{c|c|c|c|c|c}
\toprule \# of Projectors &2&5&10&15&20\\\midrule
mAP &32.2&\textbf{34.6}&32.2&32.5&32.7\\
\bottomrule
\end{tabular}
\vspace{-1em}
\end{table}

 \begin{figure}[t]
\centering
\includegraphics[width=8cm]{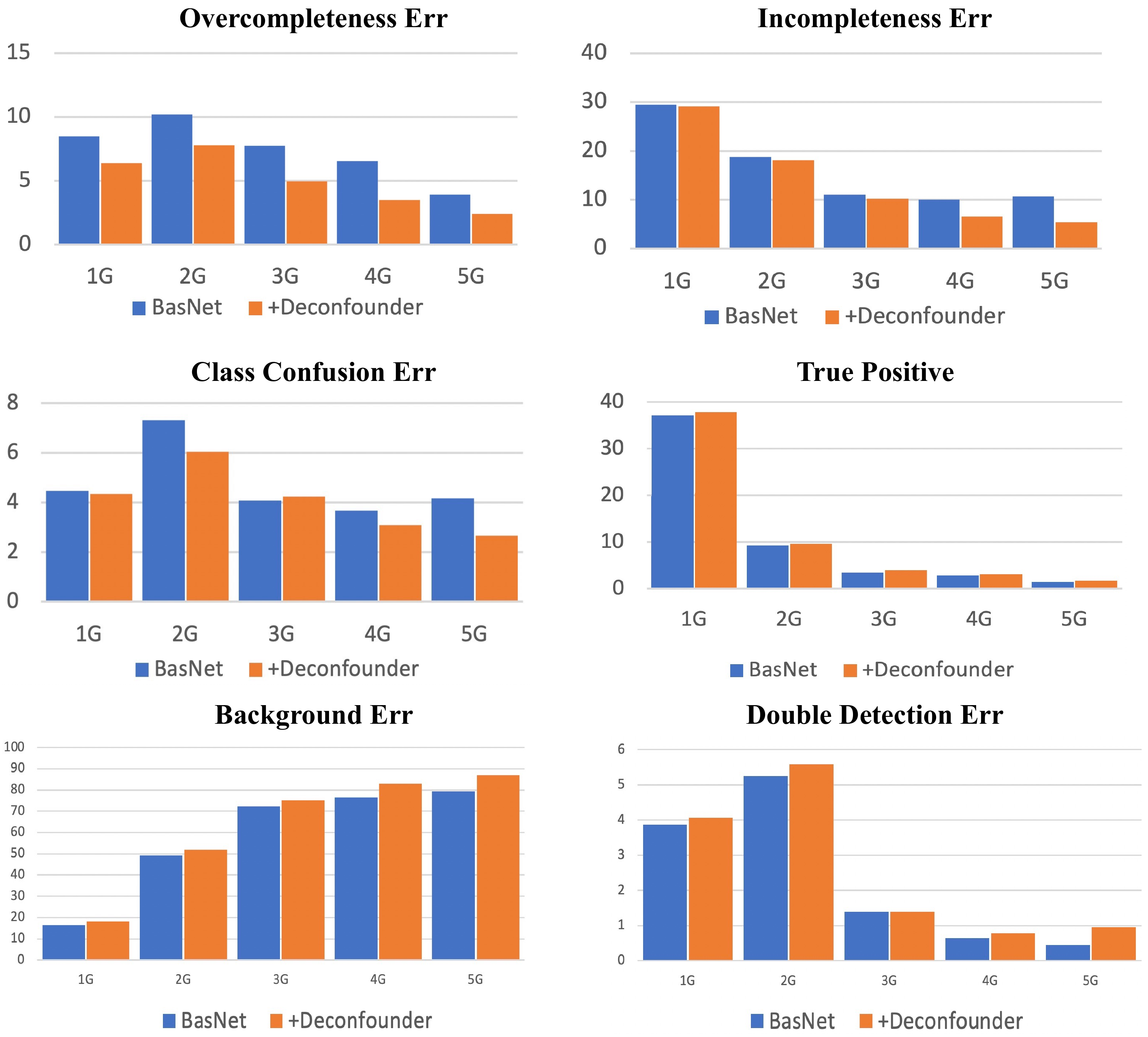}
\caption{False positive analysis on the three error types, namely \emph{Over-completeness}, \emph{Incompleteness}, and \emph{Confusion}. G is the number of ground truth instances.}
\label{fpa}
\vspace{-1em}
\end{figure}

\begin{figure*}[htb]
\label{architecture}
\centering\includegraphics[width=16cm]{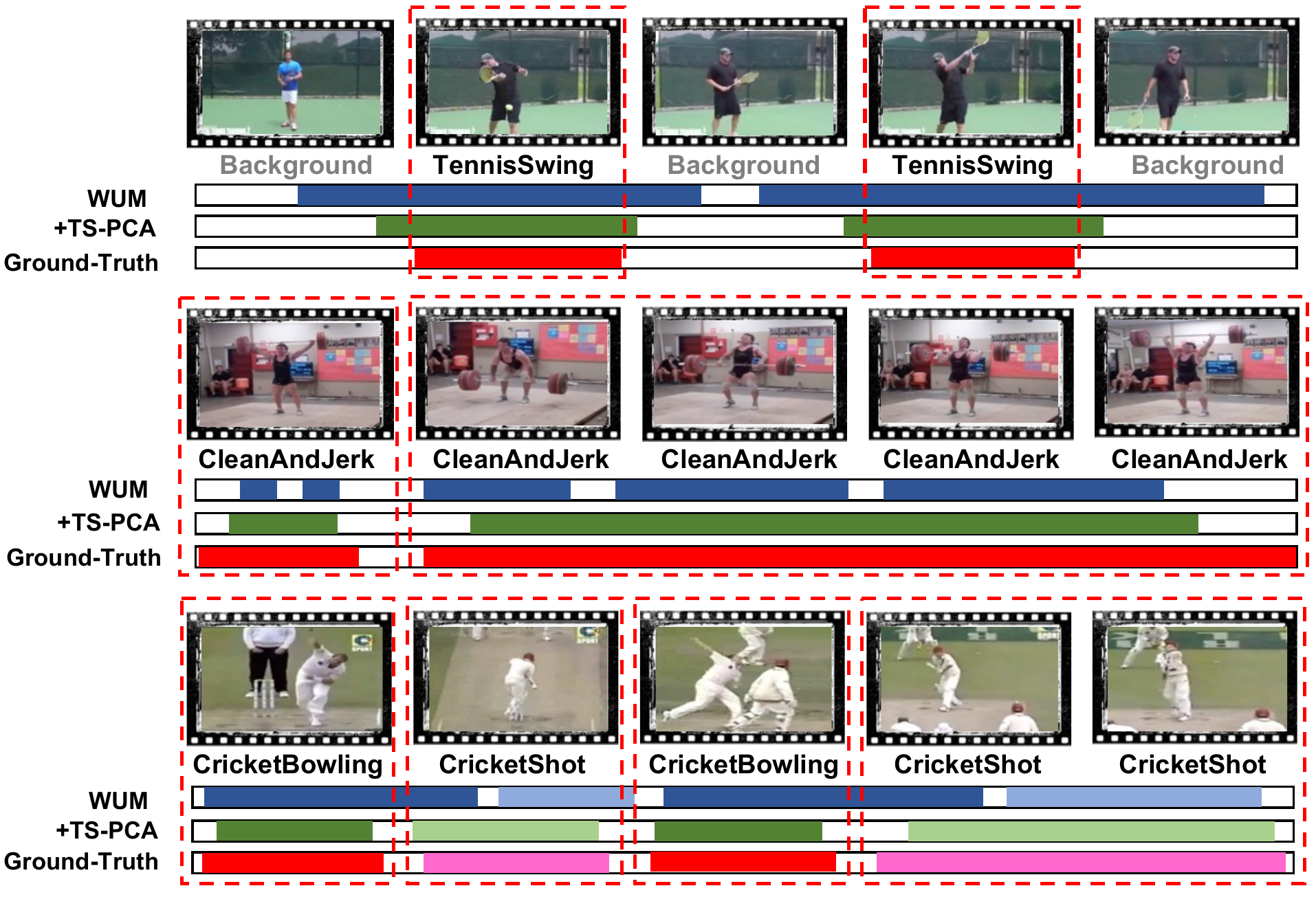}
\vspace{-1em}
\caption{Visualization of the predicted action instances on THUMOS-14. Three types of common issues: \emph{Over-completeness} (Top), \emph{Incompleteness} (Mid) and \emph{Confusion} (Bottom), are illustrated. It could be observed that, with the deconfounder to calibrate the CAS scores generated by WTAL model, more accurate action instances are localized.}
\label{example}
\vspace{-0.5em}
\end{figure*}

\subsubsection{Performance over Different Number of Feature Projectors}
Recall that TS-PCA learned $L$ feature projectors to learn $\hat{\mathbf{z}}$ and reconstruct $\mathbf{x}$. To investigate the performance of TS-PCA over different projector settings, we show the performance with respect to the number of projectors in Table~\ref{num_pro}.
The results show that projecting each input video feature onto the first $5$ principal components can preserve the data's variation to a large extent.
When we add more feature projectors, noise may be introduced. Therefore, the number of feature projectors is set to 5.

\subsection{False Positive Analysis}
\label{exp:fpa}
Following DETAD~\cite{detad}, we utilize  BasNet~\cite{basnet} on THUMOS-14 as the baseline  model  to  conduct  the false positive analysis~\cite{detad} on \emph{Over-completeness}, \emph{Incompleteness}, and \emph{Confusion}.
We make detailed analysis on the top-5G predictions, where G is the number of ground truth instances.
To observe the trend of each error type, we split the top-5G predictions
into five equal splits and investigate the percentage of each error type.
To highlight the comparisons, we reorganize the figures in Figure~\ref{fpa}: BasNet+TP obtains lower error rates in all three types of errors when compared with BasNet. As we have minimized the three error types caused by confounders, the rest of the error types that caused by the lack of full supervision become more significant compared to those in baseline. Note that the higher percentage of background error and double detection error do NOT mean that ours have more errors, because the total number of errors of ours is smaller than the baseline.

\subsection{Qualitative Results}
\label{exp:qr}
As mentioned in Section~\ref{sec:intro}, \emph{Over-completeness}, \emph{Incompleteness}, and \emph{Confusion} are the three types of localization errors frequently encountered in WTAL.
The deconfounder theory infers a substitute confounder to calibrate conventional CAS, alleviating these problems.
Figure~\ref{example} illustrates the qualitative results related to the above three common issues:
 1) \emph{Over-completeness} (Top): it's hard for the WTAL model WUM to discriminate background from foreground. With the assistance of deconfounded strategy, false-positive background segments are largely reduced. 
 2) \emph{Incompleteness} (Mid): the unobtrusive action segments are easily overlooked by WUM
 due to confounding bias. The deconfounded CAS could improve the completeness of action instances. 
 3) \emph{Confusion} (Bottom): when two or more action categories such as \textsc{CricketBowling} and \textsc{CricketShot} are visually correlated, it's hard for the WTAL model to distinguish them. The deconfounded CAS can also tackle this challenge.

%% file: text/conclusion.tex
\section{Conclusion}
In this paper, we first summarized the three fundamental localization errors in WTAL. Then, we proposed a causal inference framework to identify that the reasons are due to the unobserved confounder. To capture the elusive confounder, we presented the unsupervised TS-PCA deconfounder, which exploits the unlabelled background to model an observed substitute for the confounder, to remove the confounding effect. Moreover, with a novel decoupled training scheme, the deconfounder is model-agnostic and could support any WTAL model in a plug-and-play fashion. Significant improvement on four state-of-the-art WTAL methods demonstrates the effectiveness and generalization ability of our deconfounded WTAL.
\\
\\
\textbf{Acknowledgements.} We thank the anonymous reviewers and ACs for their valuable comments. We thank Chong Chen at Damo Academy for his valuable suggestions on TS-PCA.

%% file: text/supp.tex
This supplementary material includes additional experimental results that are not presented in the main paper and more qualitative results to demonstrate the superior performance of our proposed TS-PCA deconfounder.

\section{Visualization of CAS Scores}

\begin{figure*}[htb]
\centering\includegraphics[width=16cm]{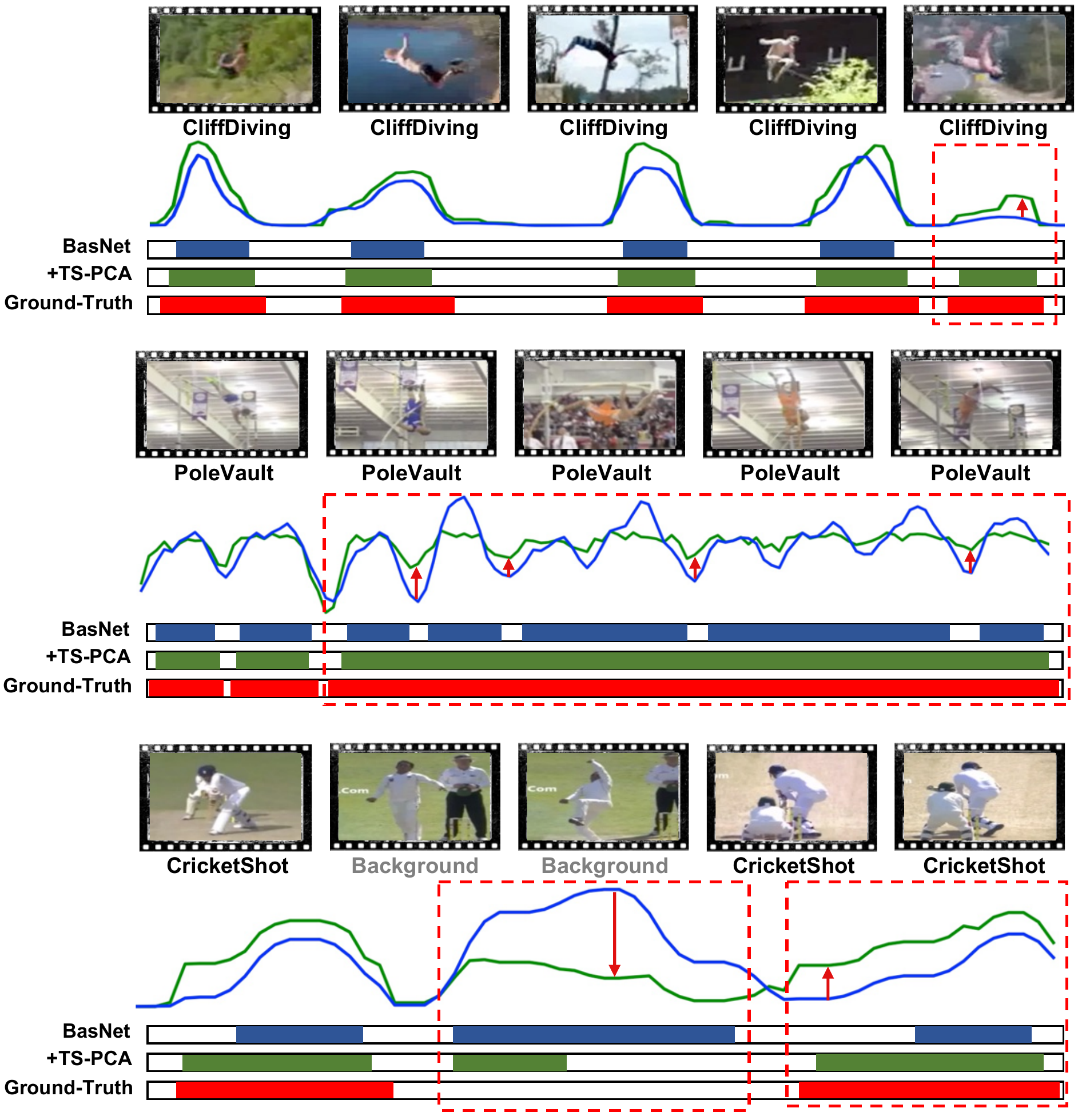}
\caption{\label{fig:cas_calibrate}
Comparisons of CASs generated by BasNet and TS-PCA on three examples from THUMOS-14. The blue line represents the CASs generated by BasNet and the green line represents the CASs generated by incorporating TS-PCA. Compared with BasNet, the CAS generated by TS-PCA can cover large and continuous regions to obtain more accurate action segments.}

\end{figure*}

To further demonstrate the effectiveness of the TS-PCA, we plot both the CAS generated by BasNet~\cite{basnet} and the CAS generated by incorporating TS-PCA (+TS-PCA) deconfounder in Figure~\ref{fig:cas_calibrate}, where the temporal CAS scores are normalized to $0$ to $1$. Compared with BasNet, the CAS generated by +TS-PCA can cover large and dense regions to obtain more accurate action segments. In the example of Figure~\ref{fig:cas_calibrate}, +TS-PCA can discover almost all the actions that are labeled in the ground-truth. However, BasNet tends to only detect the most salient regions of action segments. These results further demonstrate the effectiveness of the TS-PCA deconfounder in calibrating CAS. 
Specifically, 

\noindent\textbf{First Example}: BasNet fails to localize \textsc{CliffDiving} with short temporal durations while CAS calibrated by TS-PCA succeeds to find the last segment.
\\
\textbf{Second Example}: BasNet has difficulty in detecting action instances with long duration perfectly, where it tends to generate incomplete instances. The calibrated CAS can tackle this problem by increasing the CAS scores of false-negative segments.
\\
\textbf{Third Example}: When the background is semantically correlated with the foreground, it is difficult to well separate them. It can be observed that the calibrated CAS suppresses feedback of background to some extent. 

\section{More Qualitative Results}
\begin{figure*}[htb]
\centering\includegraphics[width=16cm]{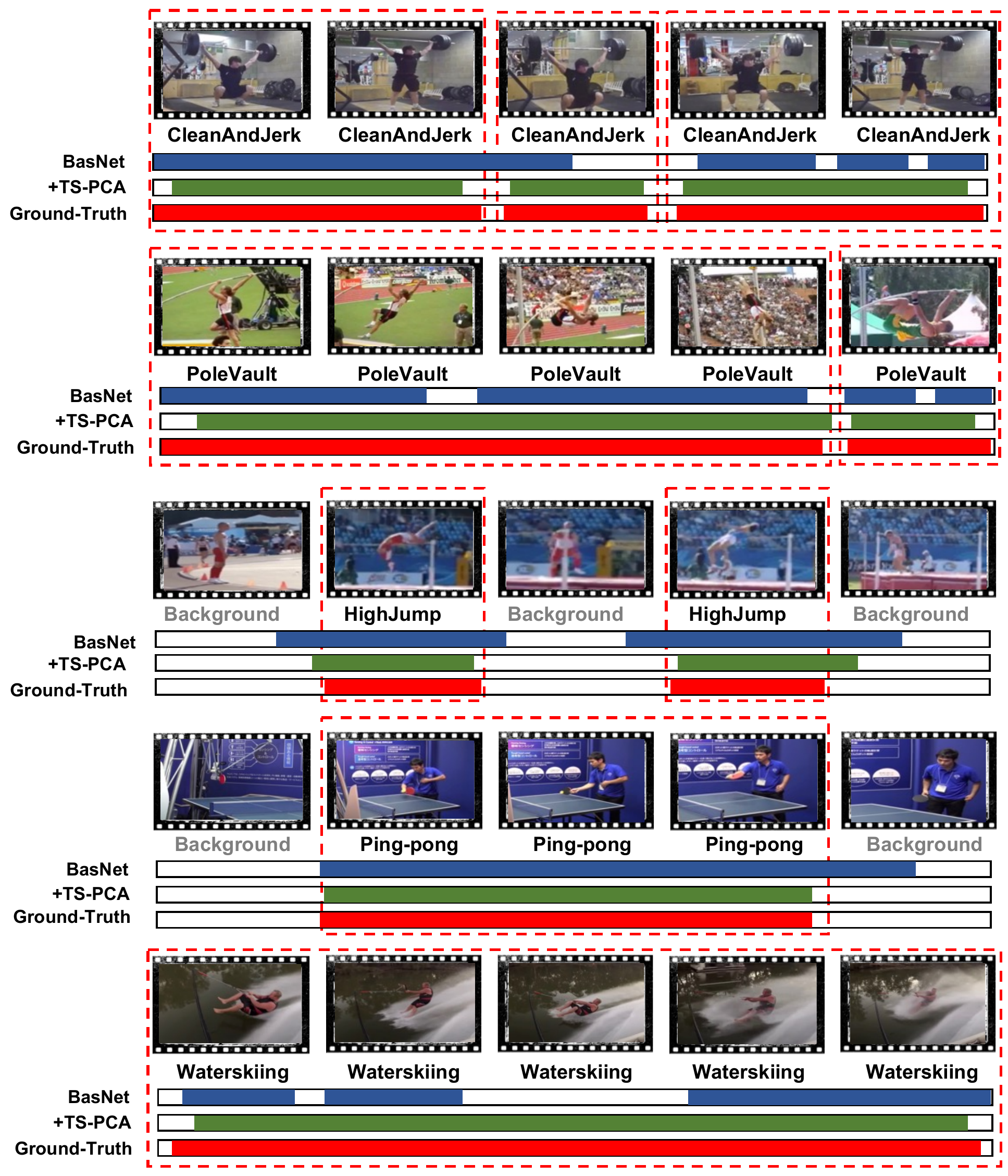}
\caption{\label{fig:supp_good}Qualitative results of BasNet with the TS-PCA deconfounder. First three rows show action instances on  THUMOS-14.
Last two rows show action instances on ActivityNet-1.3.
Red dashed rectangles highlight the improved segments predicted by +TS-PCA.}

\end{figure*}

More qualitative results are illustrated in Figure~\ref{fig:supp_good}. The first three rows are videos from the testing set of THUMOS-14 \cite{thumos} and the last two rows are from the validation set of ActivityNet-1.3 \cite{activitynet}. It can be observed that, with the TS-PCA deconfounder to calibrate CAS, more accurate action instances are localized, which demonstrates the effectiveness of the proposed deconfounder strategy.

\begin{figure*}[htb]
\centering\includegraphics[width=16.5cm]{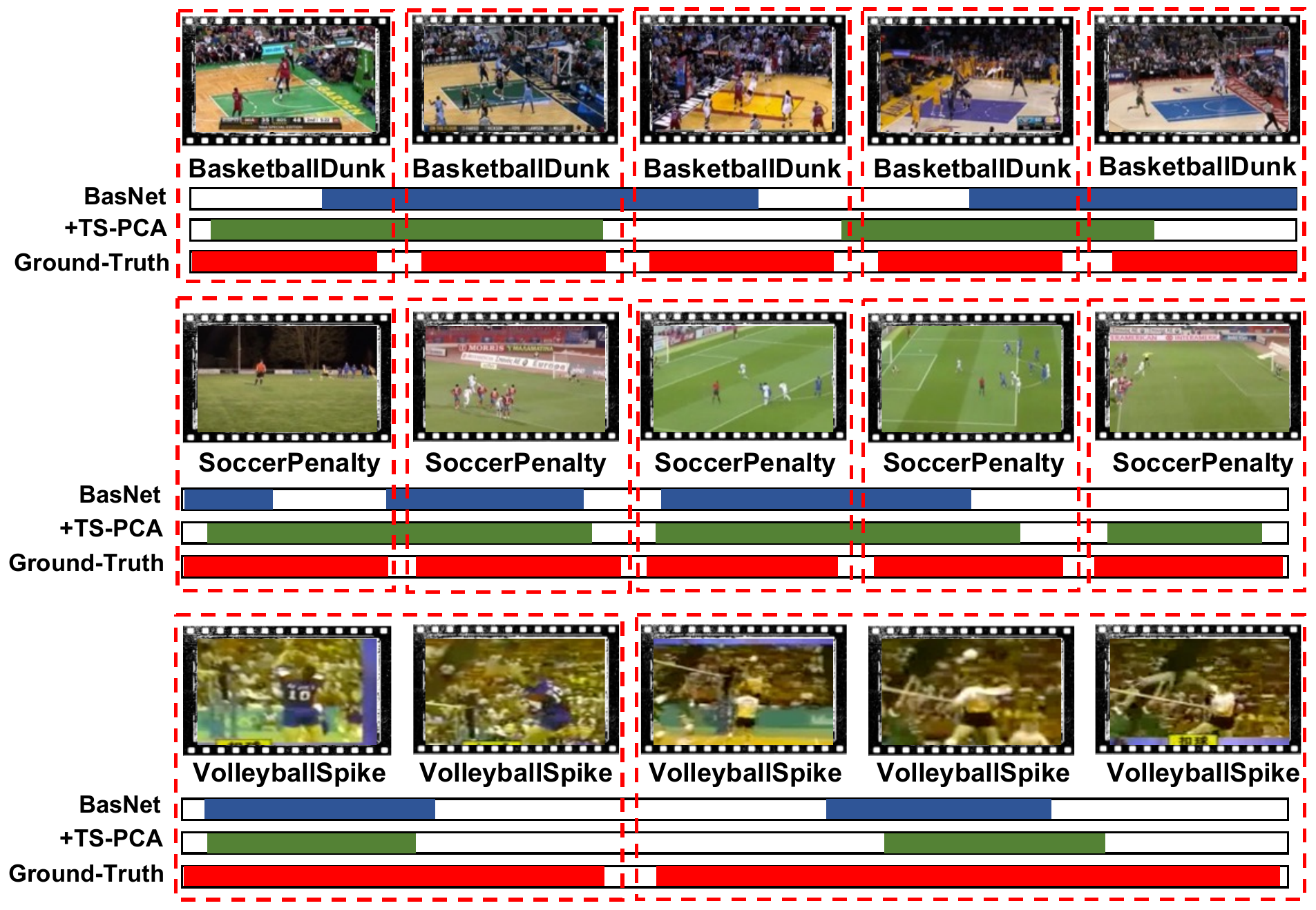}
\caption{\label{fig:supp_bad}Failure cases generated by
BasNet with the TS-PCA deconfounder on THUMOS-14.}

\end{figure*}

\section{Failure Cases}
We also show some failure cases in Figure~\ref{fig:supp_bad}. For action instances which only occupy small spatial regions of the whole frame (first two rows), it is challenging to well localize them. Moreover, if the videos are of low quality (last row), it will be hard to capture the corresponding semantic meanings, which thereby results in inaccurate proposals.